%% file: acl_latex.tex
\pdfoutput=1
\documentclass[11pt]{article}

\usepackage[preprint]{acl}

\usepackage{times}
\usepackage{latexsym}

\usepackage[T1]{fontenc}
\usepackage[utf8]{inputenc}

\usepackage{microtype}

\usepackage{inconsolata}

\usepackage{graphicx}
\usepackage{subfigure}
\usepackage{caption}
\input{settings}

\input{math}

\title{Unveiling Factual Recall Behaviors of Large Language Models through Knowledge Neurons}

\author{
    Yifei Wang\thanks{Equal contribution.}\textsuperscript{\rm 1,\rm 2 },Yuheng Chen\footnotemark[1]\textsuperscript{\rm 2 },Wanting Wen\textsuperscript{\rm 1 },Yu Sheng\textsuperscript{\rm 1,\rm 2},Linjing Li \textsuperscript{\thanks{Corresponding author.}\rm 1,\rm 2,\rm 3},    Daniel Zeng\textsuperscript{\rm 1,\rm 2}\\
    $^1$ State Key Laboratory of Multimodal Artificial Intelligence Systems, \\
    Institute of Automation, Chinese Academy of Sciences, Beijing, China \\
$^2$ School of Artificial Intelligence, University of Chinese Academy of Sciences, Beijing, China \\
$^3$ Beijing Wenge Technology Co., Ltd, Beijing, China \\
\texttt{\{wangyifei2022, chenyuheng2022\}@ia.ac.cn} \\
\texttt{\{wanting.wen, shengyu2021, linjing.li, dajun.zeng\}@ia.ac.cn}
}
\begin{document}
\maketitle
\begin{abstract}
% internal knowledge stored in the parameters of LLMs
In this paper, we investigate whether Large Language Models (LLMs) actively recall or retrieve their internal repositories of factual knowledge when faced with reasoning tasks. Through an analysis of LLMs' internal factual recall at each reasoning step via Knowledge Neurons, we reveal that LLMs fail to harness the critical factual associations under certain circumstances. Instead, they tend to opt for alternative, shortcut-like pathways to answer reasoning questions. By manually manipulating the recall process of parametric knowledge in LLMs, we demonstrate that enhancing this recall process directly improves reasoning performance whereas suppressing it leads to notable degradation. Furthermore, we assess the effect of Chain-of-Thought (CoT) prompting, a powerful technique for addressing complex reasoning tasks. Our findings indicate that CoT can intensify the recall of factual knowledge by encouraging LLMs to engage in orderly and reliable reasoning. Furthermore, we explored how contextual conflicts affect the retrieval of facts during the reasoning process to gain a comprehensive understanding of the factual recall behaviors of LLMs. 

% Code and data will be available soon.

% These discoveries provide valuable insights into the behavior of factual associations when LLMs deal with reasoning tasks, which contributes to designing effective algorithms to facilitate models to explore and apply their internal repositories of factual knowledge more comprehensively and thereby improve the overall reliability of reasoning. 
\end{abstract}

\section{Introduction}
\begin{figure}[!t]
    \centering
    \includegraphics[width=1.0\linewidth]{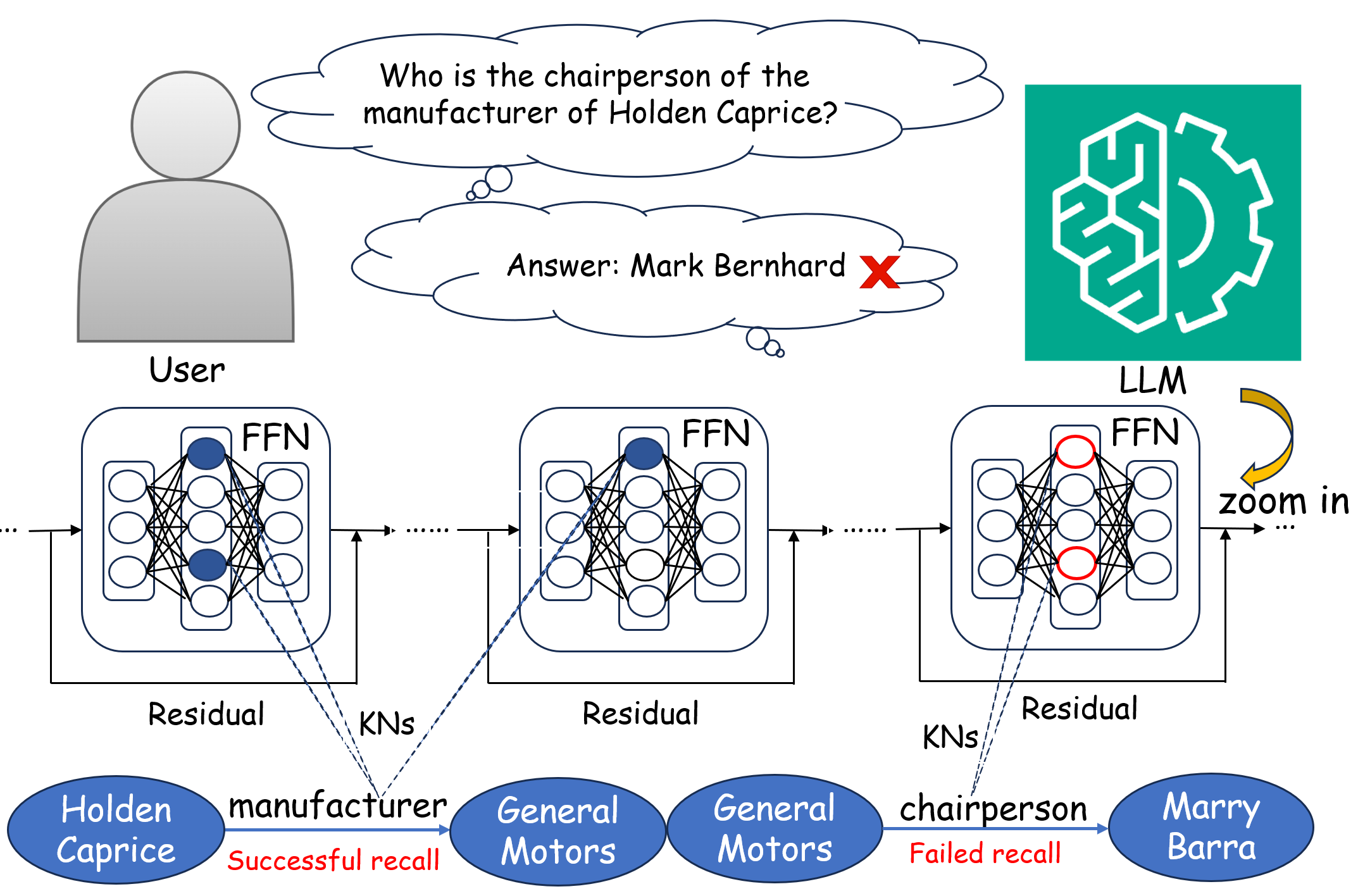}
    \caption{An unsuccessful case of reasoning due to factual retrieval failure of the triplet (General Motors, chairperson, Marry Barra).}
    \vspace{-.4cm}
    \label{fig1}
\end{figure}

Recent advancements in Large Language Models have underscored their exceptional \emph{reasoning} prowess with natural language understanding across a broad spectrum of tasks \cite{chen2023program, NEURIPS2022_8bb0d291, NEURIPS2020_1457c0d6, creswell2023selectioninference}. However, amidst these achievements, a specific form of reasoning has been somewhat overlooked and insufficiently investigated: reasoning tasks that require the utilization of internal factual knowledge associations. For instance, when presented with a 2-hop question such as "Who is the chairperson of the manufacturer of the Holden Caprice?" in Figure \ref{fig1}, LLMs must first identify that the manufacturer of the Holden Caprice is General Motors, and subsequently retrieve the name of General Motors' chairperson from their internal knowledge, also referred to as parametric knowledge \cite{neeman-etal-2023-disentqa,zhong2024seeking}. Previous work has shown factual knowledge emerges in both GPT \cite{meng2022locating} and Bert models \cite{petroni-etal-2019-language,10.1162/tacl_a_00324}. Unlike mathematical \cite{10.1093/oxfordhb/9780195325928.003.0004} and logical reasoning \cite{pan-etal-2023-logic}, factual reasoning heavily relies on the factual knowledge encoded within LLMs, acquired through extensive pretraining on vast corpora, rather than on user-inputted premises. At the same time, it differs from commonsense reasoning \cite{zhao2023large,trinh2019simple}, which taps into general knowledge acquired through dynamic training to foster a holistic understanding of the world, instead of emphasizing specific factual information. 
% \footnote{\url{http://www- formal.stanford.edu/leora/commonsense/}}

Intuitively, it is reasonable to expect LLMs to harness extensive parametric knowledge to tackle reasoning tasks. Yet, an important question emerges: How effectively can LLMs actually retrieve and utilize their internal knowledge for reasoning purposes? Delving into this question is crucial for several reasons. First, efficient use of parametric knowledge may significantly reduce reliance on external data sources, thereby lowering operational costs of data retrieval and API usage. Second, this dynamic capability allows the knowledge within LLMs to flow and interconnect \cite{onoe-etal-2023-lms}, showcasing these models as organic entities rather than static information repositories \cite{petroni-etal-2019-language}. From a practical perspective, the accurate retrieval and application of parametric knowledge lead to more reliable and interpretable reasoning, enhancing their utility and trustworthiness in real-world applications.
% Firstly, if LLMs can efficiently access the stored knowledge, it reduces greatly reliance on external data sources, thereby lowering operational costs associated with data retrieval and API usage. Secondly, it implies that the factual knowledge encoded within LLMs possesses a dynamic capacity to flow and interconnect\cite{onoe-etal-2023-lms}, showcasing these models as organic integrities rather than static repositories of information\cite{petroni-etal-2019-language}. From a practical perspective, the ability of LLMs to accurately retrieve and apply relevant facts would lead to more reliable and interpretable reasoning outcomes, enhancing the overall utility and trustworthiness of these models in real-world applications. 

Transformer-based language models have accumulated substantial knowledge through extensive pretraining \cite{NIPS2017_3f5ee243}. A significant body of recent research has focused on the factuality issues of LLMs \cite{wang2023survey}. One stream of this research has concentrated on pinpointing the locations within these models' architectures where factual knowledge is stored and encoded \cite{meng2022locating, dai-etal-2022-knowledge, singh-etal-2020-bertnesia, geva-etal-2022-transformer, geva-etal-2021-transformer}. Simultaneously, there has been a concerted effort to understand the mechanism by which this knowledge is \emph{accessed} during the inference phase \cite{geva-etal-2023-dissecting, yang2024large}. Another line of work discusses the balance of the retrieved knowledge and its parametric counterparts \cite{kwiatkowski-etal-2019-natural, Kandpal2022LargeLM,yu2023generate}. However, the majority of these studies have either been confined to elementary retrieval tasks, such as recalling a single fact object $o$ from a given triplet $(s, r, o)$, or have not delved into the intricacies of factual knowledge recall and utilization in more advanced challenges, particularly within complex reasoning scenarios. Our work addresses these limitations by examining the inner dynamics of factual recall within LLMs during the two-hop factual reasoning process, providing fresh insights into the behavior of factual recall in reasoning and highlighting avenues for enhancing the robustness and reliability of reasoning through more sophisticated knowledge utilization strategies.

In this work, we investigate the harness of internal knowledge for reasoning through the lens of Knowledge Neurons (KNs). We focus on the basic setting of factual reasoning involving the composition of two facts (for example, "Who is the chairperson of the manufacturer of Holden Caprice?" in Figure \ref{fig1}). To achieve this, we carefully craft two-hop reasoning questions dataset that seamlessly integrates with the KN technique. We assess the level of factual recall at each reasoning step by introducing a novel metric, KN Scores. We examine KN Scores under three conditions of two-hop reasoning: no CoT, zero-shot CoT, and few-shot CoT, unveiling the pitfalls existing in the reasoning process and the enhancement effect of CoT \cite{NEURIPS2022_9d560961}. 
Then we conduct targeted interventions on KNs to enhance or suppress the factual retrieval process, finding the contributing impact on reasoning performance. Furthermore, we provide a detailed analysis of factual shortcuts \cite{ju2024investigating,10.1145/3596490,li-etal-2024-understanding}, potentially caused by redundant information stored in models' parameters within LLMs used for reasoning.  Finally, we explore how the presence of knowledge conflict outside LLMs influences the factual recall process.
% To study the behavior of factual retrieval in reasoning tasks for LLMs, we experiment with instructed versions of LLaMA2-7B \cite{touvron2023llama}, LLaMA-3-8B, Mistral-7B \cite{jiang2023mistral}.
% \footnote{\url{https://huggingface.co/meta-llama/Meta-Llama-3-8B-Instruct}}
Our findings can be summarized as follows:
\begin{itemize}[itemsep=1pt, parsep=0pt, topsep=0pt,leftmargin=15pt]
    \item LLMs do not consistently retrieve the pertinent factual knowledge essential for reasoning, with more than a third of reasoning errors stemming from deficiencies in the retrieval of factual associations.
    \item  CoT could remarkably enhance the recall of factual knowledge by facilitating engagement in step-by-step reasoning, thereby reducing the likelihood of shortcuts.
    \item By enhancing and suppressing the recall process, we demonstrate that successful factual retrieval is a pivotal factor in improving reasoning performance.
    \item The presence of knowledge conflict in context could enhance the retrieval of the corresponding fact in the reasoning process to a degree.
\end{itemize}
% Our findings can be summarized as follows: (1) LLMs do not always retrieve the relevant factual knowledge necessary for reasoning, and 37 percent of reasoning failures are attributed to issues with the retrieval of factual associations. (2) We provide evidence for the existence of shortcuts, showing that some reasoning questions can be successfully answered even if one or both of the facts involved in the reasoning tasks are not utilized and recalled. (3) By enhancing and suppressing the recall process, we demonstrate that the extent of factual recall positively correlates with reasoning accuracy. (4) CoT could remarkably improve the recall of factual knowledge, facilitating LLMs' engagement in mediating tasks and greatly enhancing successful reasoning outcomes.

\section{Preliminaries}
\subsection{Problem Formulation}
We represent facts, such as "(Holden Caprice, manufacturer, General Motors)", as a triplet $(s, r, o)$, where $s$ is the subject, $r$ is the relation, and $o$ is the object. We formulate two-hop factual reasoning questions as a composition of two linked facts $((s, r_1, o_1),(o_1, r_2, o_2))$, with a bridge entity $o_1$ connecting them. To query LLMs, these triplets must be converted into natural language queries. For a single relation $r$, we instruct ChatGPT (gpt-3.5-turbo) to generate query templates as $QT_r(\cdot)$. For instance, the single-relation triplet (Holden Caprice, manufacturer, General Motors) can be converted as $QT_{manufacturer}(Holden Caprice)$: "Which company manufactures Holden Caprice?". Similarly, for a composition of two relations $r_1$ and $r_2$, we prompt ChatGPT to generate a query template as $QT_{r_2}(r_1(\cdot))$, with $r_1(\cdot)$ denoting the description of the entity related to $s$ via $r_1$ relation (e.g. The manufacturer of Holden Caprice). We refer to the single-hop query as $QT_{1H}$ and the two-hop query as $QT_{2H}$. 
 
 We consider an autoregressive language model $F: X\rightarrow Y$, which accepts an input $x\in X$ and produces a prediction $y\in Y$, continuing the input $x$. We deem that the model "knows" a fact $(s, r, o)$ if the output $F(QT_{r}(s))$ matches the ground label $o$ and that LLMs can reason a question involving two-hop fact triplets $((s, r_1, o_1),(o_1, r_2, o_2))$ successfully if the output $F(QT_{r_2}(r_1(s))$ matches the ground label $o_2$. It is noteworthy that query templates, even for the same single relation, are generated with diversity by ChatGPT. This diversity discourages models from making predictions based on the occurrence of specific words, ensuring that they recall knowledge from within themselves instead. We denote the set of two-hop factual questions as $\Omega$, with $\Omega_{T}$ representing the subset of questions that LLMs can answer correctly and $\Omega_F$  denoting the subset of questions that LLMs cannot answer correctly. For simplicity, we use $\zeta$ to denote $((s, r_1, o_1), (o_1, r_2, o_2))$, thus we have:
 \begin{equation}
\begin{minipage}{0.88\linewidth}
\resizebox{\linewidth}{!}{$\displaystyle
\Omega_{T} = \left\{\zeta \mid \operatorname{F}_{{\theta}}(\operatorname{QT}_{r_2}(r_1(s))) = o_2, \forall \zeta \in \Omega \right\}
$}
\end{minipage}
\end{equation}
\vspace{-.5cm}
\begin{equation}
\begin{minipage}{0.88\linewidth}
\resizebox{\linewidth}{!}{$\displaystyle
\Omega_{F} = \left\{\zeta \mid \operatorname{F}_{{\theta}}(\operatorname{QT}_{r_2}(r_1(s))) \ne o_2, \right. \left. \forall \zeta \in \Omega \right\}
$}
\end{minipage}
\end{equation}
%  {\small
%  \vspace{-.3cm}
%  \begin{align}
% &\Omega_{T} = \left\{\zeta \mid F_{{\theta}}(QT_{r_2}(r_1(s))) = o_2, \right.  \left. \forall \zeta \in \Omega \right\} \\
% &\Omega_{F} = \left\{\zeta \mid F_{{\theta}}(QT_{r_2}(r_1(s))) \ne o_2, \right. \left. \forall \zeta \in \Omega \right\}
% \end{align}
%  }
\subsection{Knowledge Neurons}
% \begin{figure}[h]
%     \centering
%     \includegraphics[width=0.7\textwidth]{figure/kn.pdf}
%     \caption{the concept of KN for a fact triplet $(s,r,o)$}
%     \label{fig:enter-label}
% \end{figure}
Pretrained language models store vast amounts of factual knowledge and have a strong ability to recall this factual knowledge without further training \cite{petroni-etal-2019-language,10.1162/tacl_a_00324}. Drawing inspiration from the key-value-memory nature of feed-forward layers \cite{geva-etal-2021-transformer}, \citet{dai-etal-2022-knowledge} proposes that factual knowledge is stored in specific neurons within the Feed-Forward Networks (FFNs) of the Transformer models, termed as knowledge neurons. They find that knowledge neurons are activated by knowledge-expressing prompts. The higher the activation of these knowledge neurons is, the more significantly their corresponding facts are expressed. Therefore, to assess the recall and utilization of the fact triplet $(s, r, o)$ necessary in the reasoning process, we refer to the activity of KNs as an indicator of factual recall. We make the following \textbf{invariant assumptions}: the KNs responsible for the expression of particular relational facts remain consistent across different application contexts. A specific fact is indicated by the same set of KNs under both single-hop queries and reasoning queries, which is a cornerstone for subsequent experiments. In Appendix \ref{B}, We detail a methodology that utilizes integrated gradient \cite{sundararajan2017axiomatic} method to compute the contribution of all neurons in the intermediate layers of FFNs to the correct prediction of a multi-token ground truth, identifying neurons with greater contributions as KNs.
\section{TFRKN: Two-hop Factual Reasoning for Knowledge Neurons}
To investigate the behavior of factual recall in reasoning tasks for LLMs, we have developed a specialized dataset for knowledge neurons called \textbf{T}wo-hop \textbf{F}actual \textbf{R}easoning for \textbf{K}nowledge \textbf{N}eurons, TFRKN.
\paragraph{Dataset Construction}  Our dataset consists of two-hop factual questions, where each question involves two facts that are connected by an intermediate entity. LLMs are more likely to recall triplets related to popular entities\cite{mallen-etal-2023-trust}. Therefore, for entity selection, we use the cumulative pageview count over the past 12 months as a metric and select the top 500 popular entities from Wikidata \cite{vrandevcic2014wikidata} based on this criterion. Two-hop fact triplets are then extracted from sub-graphs consisting solely of a set of manually selected relations and entities. To identify KNs for each-hop fact, we reformulate each fact triplet into more than five varied natural questions using ChatGPT (Appendix \ref{A}). The TFRKN dataset encompasses 4,550 distinct instances covering 213 unique relational combinations with a sample instance shown in Table \ref{table6}. 

\section{Diagnose the Pitfalls of Factual Recall in Reasoning}
In the realm of two-hop factual reasoning, an optimal and dependable reasoning trajectory is a multi-hop reasoning approach \cite{Welbl2017ConstructingDF,ju2024investigating}. This process requires identifying the bridge entity first and then using it to solve the second hop question, necessitating that LLMs recall the relevant fact at each hop step by step, culminating in the formulation of the correct answers. In this section, we investigate whether LLMs faithfully retrieve factual knowledge at each hop when undertaking reasoning tasks.
% To evaluate the efficacy of factual recall within LLMs during reasoning tasks, we devise a novel metric termed KN Scores, which quantifies the LLMs' capacity for the internal recall of a specific fact during the reasoning process. Then we demonstrate the extent to which LLMs are capable of recalling factual knowledge incrementally throughout the reasoning and elucidate the impact that the CoT has on this recalling process. Ultimately, our assertions are corroborated by a comprehensive suite of experiments conducted on the TFRKN dataset. 
\subsection{KN Scores}
To evaluate the efficacy of factual recall within LLMs during reasoning tasks, we devise a novel metric termed KN Scores as follows:
% To quantify the capacity for internal recall of specific facts within LLMs, we devise a novel metric, termed KN Scores, as follows:
\begin{align}
    &\operatorname{FFN}^{(l)}(\operatorname{H}^{(l)}) = \operatorname{W}_{2}^{(l)}\operatorname{SiLU}(\operatorname{H}^{(l)}\operatorname{W}_{1}^{(l)}) \\
    &\omega_i^l = \operatorname{SiLU}(\operatorname{H}^{(l)}\operatorname{W}_{1}^{(l)})[i], \quad \forall \omega_i^l \in \omega \\[-3pt]
    &\operatorname{KN\ Scores} = \frac{1}{|\omega|} \sum \omega_i^l, \forall \omega_i^l \in \omega
\end{align}
where $\operatorname{H}^{(l)}$ represents the input to the FFN of the $l$-th layer, which consists of the outputs from the $l$-th attention layer combined with the residual stream; $\omega_i^l$ denotes the $i$-th neuron in the $l$-th intermediate layer of FFN; $\omega$ represents the KNs associated with a specific fact triplet, denoted as $(s, r, o)$; $|\omega|$ denotes the size of the set, i.e., the number of KNs; and $\operatorname{SiLU}$ denotes the activation function. For the first-hop and second-hop fact, we designate their respective sets of KNs as $\omega_{1}$ and $\omega_{2}$. Under the context of a single-hop query, we denote KN Scores as $\{\overline{\omega}|QT_{1H}\}$. Similarly, within the two-hop reasoning context, KN Scores are represented as $\{\overline{\omega}|QT_{2H}\}$.
\begin{figure}[!t]
    \centering
    \includegraphics[width=1\linewidth]{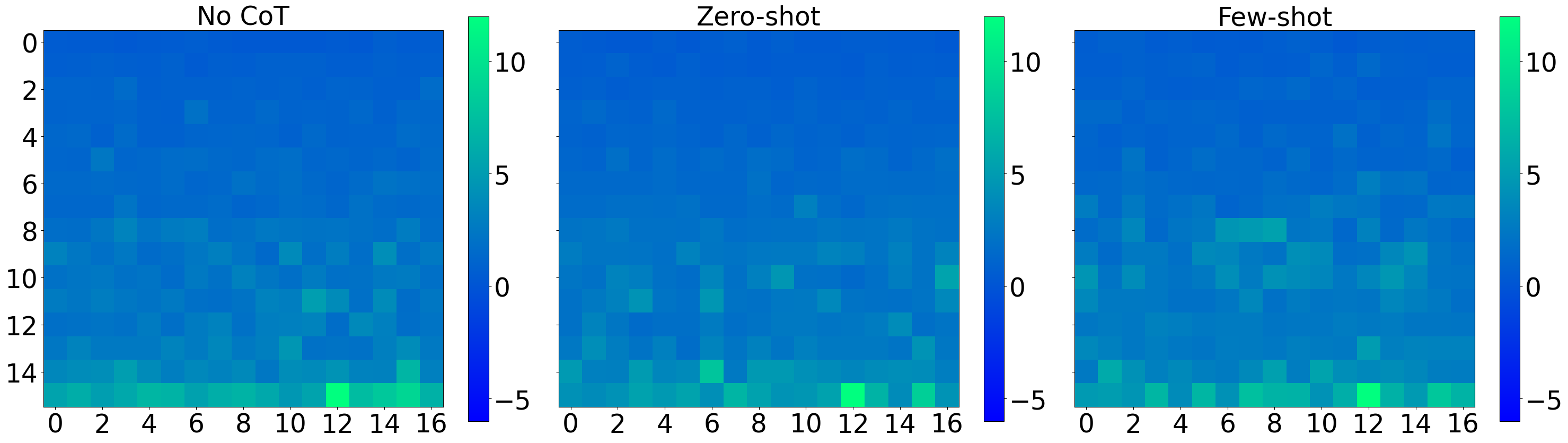}
    \caption{Scaled visualization of neuron activities within the intermediate layers of FFNs in Mistral-7B for the same case (A 32-layer$\times$14336-neuron matrix). The vertical axis shows the depth of layers, while the horizontal axis shows the neuron index in the FFN's intermediate layers. It is evident that KNs are distributed in the middle and final layers.}
    \label{fig2}
\end{figure}
\subsection{Experiment}
\paragraph{Setup} We begin by filtering out reasoning questions where LLMs are unable to recall all individual facts, ensuring that any reasoning failures are due to the models' inability to retrieve factual information rather than a lack of the foundational knowledge necessary for performing reasoning tasks. We then proceed to employ $\text{Fact}_1 \text{Query}$ and $\text{Fact}_2 \text{Query}$ (in Table \ref{table6}) from each data point to pinpoint the positions of KNs for each-hop fact. Then we hook the values of each neuron belonging to $\omega_{1}$ and $\omega_{2}$ across various query scenarios to compute KN Scores. Using the KN Scores metric, we evaluate the recall of each fact under three distinct experimental conditions:
\textbf{no CoT}, \textbf{zero-shot CoT}, and \textbf{few-shot CoT}. For each condition, we record KN Scores for both the first-hop $\{\overline{\omega}_{1}|QT_{2H}\}$ and the second-hop $\{\overline{\omega}_{2}|QT_{2H}\}$ facts within the context of two-hop reasoning questions. We select the KN Scores $\{\overline{\omega}_{1}|QT_{1H}\}$  and $\{\overline{\omega}_{2}|QT_{1H}\}$ under single-hop queries as baselines since KNs are significantly active in that straightforward context. We experiment with the instructed versions of three popular open-source models: LLaMA2-7B \cite{touvron2023llama}, LLaMA3-8B, Mistral-7B \cite{jiang2023mistral} (see Appendix \ref{C} for more experimental details).
\subsection{Results}
\paragraph{Single-hop vs. Muti-hop Reasoning} In reasoning scenarios, LLMs access their internal knowledge less frequently in comparison to the straightforward retrieval of single-hop facts. Table \ref{table1} illustrates a notable decrease in KN Scores for all single-hop facts when addressing two-hop reasoning questions. This observation strongly indicates that, in reasoning contexts, LLMs tend to either fail to recall the bridge entity or struggle to identify the second-hop relation, leading to the failure of executing the remaining multi-hop reasoning as anticipated. Compared to directly recalling single-hop facts (e.g., "Who is the chairperson of General Motors?"), it is more challenging for LLMs to recall and organize relevant facts for reasoning. LLMs may take alternative salient pathways existing in their parameters, such as shortcuts, rather than engaging in systematic, step-by-step reasoning.
% \emph{In reasoning scenarios, the frequency with which LLMs access their internal knowledge diminishes markedly compared to just recalling single-hop fact.} Table 1 illustrates a notable decrease in the activation levels of KNs for all single-hop facts under two-hop reasoning questions across the experimental LLMs, which strongly suggests that, in certain reasoning contexts, LLMs either fail to recall the bridge entity or struggle to identify the second-hop relation, which both consequently leads to the failure of executing the remaining multi-hop reasoning as anticipated. This strongly demonstrates that, compared to directly recalling single-hop facts ("Who is the chairperson of General Motors?"), it is more challenging for LLMs to recall and organize these relevant facts for reasoning. In other words, LLMs may take alternative salient pathways existing in their parameters such as shortcuts rather than engaging in systematic, step-by-step reasoning. 
% Taking Mistral-7B as an example, the KN Score for the first-hop fact decreases by 10.84 percent, while the activation for the second-hop fact drops by 11.77 percent.
% \vspace{-.17cm}
\begin{figure*}[!ht]
    \centering
    \subfigure{\includegraphics[width=.43\linewidth]{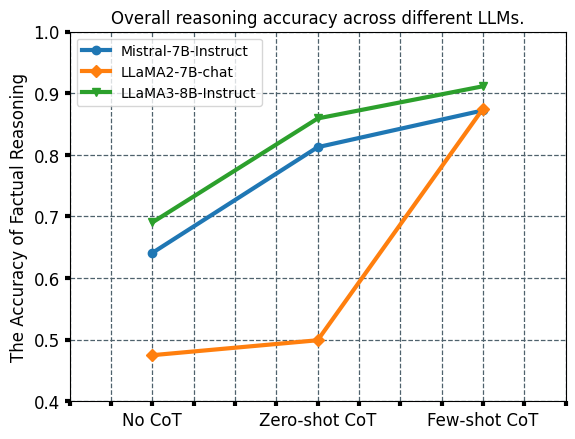}}
    \hfill
    \subfigure{\includegraphics[width=.55\linewidth]{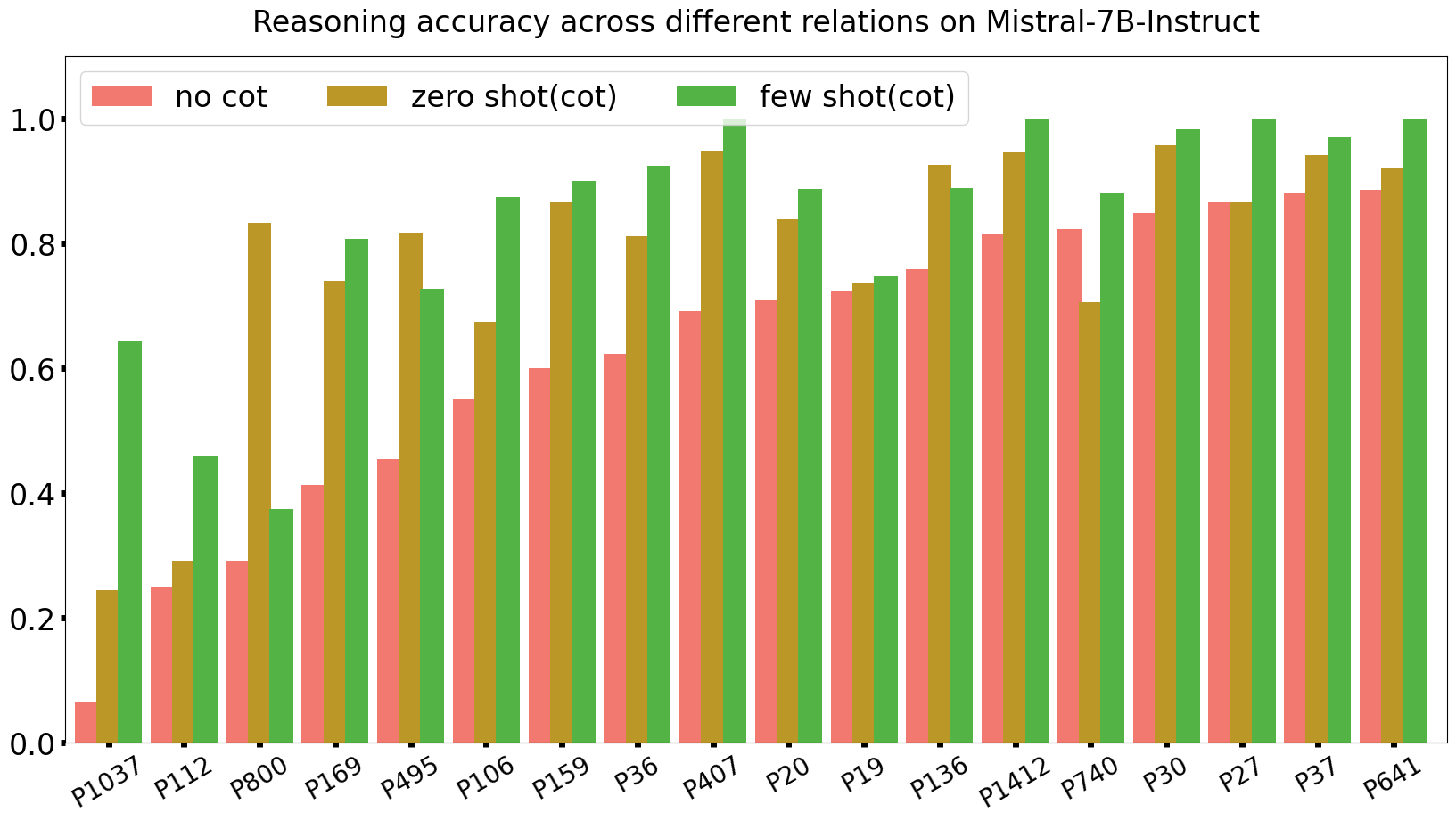}}
    \caption{Overall reasoning performance on TFRKN under different CoT situations.}
    \label{fig3}
    % \vspace{-.4cm}
\end{figure*}
\begin{table}[t]
    \centering
    \small
    \setlength{\tabcolsep}{3pt}
    % \captionsetup{belowskip=0pt,aboveskip=0pt}
    \begin{tabular}{l c c c c c c}
    \toprule
    \textbf{Models} & \multicolumn{2}{c}{\textbf{Mistral-7B}} & \multicolumn{2}{c}{\textbf{LLaMA2-7B}} & \multicolumn{2}{c}{\textbf{LLaMA3-8B}} \\
    \cmidrule{2-7}
    & \textbf{$\overline{\omega}_{1}$} & \textbf{$\overline{\omega}_{2}$} & \textbf{$\overline{\omega}_{1}$} & \textbf{$\overline{\omega}_{2}$}& \textbf{$\omega_{1}$} & \textbf{$\overline{\omega}_{2}$} \\
    \cmidrule{2-7}
    \cmidrule{2-7}
    \cmidrule{2-7}
    % \cmidrule{lr}{2-3} \cmidrule{lr}{4-5} \cmidrule{lr}{6-7}
    Single-hop & 2.44  & 2.61  & 2.01 & 1.89 & 1.70 & 1.72 \\
    \midrule
     & \textbf{$\Delta_{\overline{\omega}_1}$}& \textbf{$\Delta_{\overline{\omega}_2}$} & \textbf{$\Delta_{\overline{\omega}_1}$}& \textbf{$\Delta_{\overline{\omega}_2}$} & \textbf{$\Delta_{\overline{\omega}_1}$}& \textbf{$\Delta_{\overline{\omega}_2}$} \\
    \cmidrule{2-7}
    \cmidrule{2-7}
    \cmidrule{2-7}
    No CoT & -10.84  & -11.77  & -13.18 & -8.18 & -10.79& -8.96 \\
    Zero-shot& 11.56  & -8.48 & -2.49 & -8.30 & 11.19 & 6.24 \\
    Few-shot& 17.36  & 2.42 & 1.32 & 2.46 & 13.00 & 7.31 \\ 
    \bottomrule
    \end{tabular}
    \caption{KN Scores for three conditions across three models. $\overline{\omega}$ is the KN Score of a specific fact while $\Delta$ indicates the change ratio (in percentages) of values compared with the single-hop baselines.}
    \label{table1}
\end{table}
\paragraph{CoT vs. No CoT} CoT, whether zero-shot or few-shot, markedly improves factual knowledge utilization in LLMs over no CoT (KNs are more activated under CoT settings in Figure \ref{fig2}), which is evidenced by a higher $\Delta_{\overline{\omega}_1}$ and $\Delta_{\overline{\omega}_2}$ compared with no CoT setting, as shown in Table \ref{table1}. We posit that \textbf{this enhancement is likely driven by the step-by-step thinking process, which further stimulates the recall of facts as multi-hop reasoning progresses.} This hypothesis can be supported by comparing the zero-shot and few-shot CoT settings. Across three models, it is clear that zero-shot CoT struggles to significantly improve the recall of the second-hop fact compared to the reinforcement of the first-hop fact recall. However, consistent improvement across both triplets can be observed for few-shot settings. This observation strongly suggests that the reasoning direction in zero-shot scenarios is unclear, which prevents models from effectively identifying which relations of facts concerning the bridge entity to retrieve. In stark contrast, few-shot scenarios often mitigate this issue. Through the acquisition of knowledge from contextual demonstrations, models are more inclined to determine the subsequent phase in the reasoning trajectory and, in turn, adeptly utilize the relevant factual information via their attention mechanisms.
\paragraph{Factual Recall vs. Reasoning Accuracy} The combination of Figure \ref{fig3} and Table \ref{fig1} illustrates a positive correlation between the recall of relevant fact triplets and reasoning accuracy. This relationship is especially pronounced in the case of LLaMA3-8B model under few-shot CoT, where the maximum increase in the recall of both $\Delta_{\overline{\omega}_1}$ and $\Delta_{\overline{\omega}_2}$ leads to the highest reasoning accuracy. However, the eliciting effect of CoT on factual recall across various LLMs is not uniform. For instance, zero-shot CoT mitigates the forgetting of factual information to some extent for LLaMA2-7B, whereas for LLaMA3-8B, zero-shot CoT enhances the retrieval of factual information to a level comparable to few-shot CoT. This adequately illustrates that the efficacy of CoT is also contingent upon the intrinsic capabilities of the LLMs themselves. 
% when they are of nearly the same scale.
% Figure 3 shows that CoT notably improves reasoning accuracy across different LLMs and relational contexts, with few-shot CoT achieving the highest overall accuracy.
\section{Interventions on the Recall of Facts}
% Section 4 gives some clues that a high recall of factual associations enhances the reliability of step-by-step reasoning, thereby leading to greater reasoning accuracy. In this section, we conduct targeted interventions on factual recall within the reasoning process to determine whether augmenting the recall of factual information improves reasoning performance. Specifically, we manually intervene the expression of particular knowledge in LLM by modifying the activation levels of KNs, thereby influencing the factual recall process.
% Section 4 provides evidence suggesting that a high recall rate of factual associations can enhance the reliability of step-by-step reasoning, which in turn leads to improved reasoning performance. In this section, we investigate the impact of factual recall on reasoning performance by conducting targeted interventions within the reasoning process. 
\subsection{Enhance and Suppress KNs}
To gain a deeper understanding of factual recall behaviors, we intervene in the retrieval of specific knowledge within LLMs by manually adjusting the activation levels of KNs. Specifically for each factual triplet \((s, r, o)\), we modulate the internal recall by adjusting the values of the KNs associated with this triplet, either amplifying or diminishing them according to Equation \ref{equation6}. 
% , we intend to selectively enhance or suppress the recall of one or both of these triplets for an in-depth analysis.
\begin{equation}
\left\{
\begin{aligned}
    & \operatorname{Enhance:} \omega_i^l = n \times \omega_i^l,  n > 1, \forall \omega_i^l \in \omega \\
    & \operatorname{Suppress:} \omega_i^l = 0, \quad \omega_i^l \in \omega
\end{aligned}
\right.
\label{equation6}
\end{equation}
\subsection{Experiment}
\paragraph{Setup} We have meticulously designed four sets of controlled experiments on TFRKN to monitor changes in reasoning outcomes. The experimental paradigms are as follows: (1) Base: We allow LLMs to respond to two-hop questions under standard conditions (2) Enhance: For questions answered incorrectly under Base situation, we amplify the activation level of KNs and subsequently assess the reasoning accuracy. (3) Suppress: Conversely, for two-hop questions correctly answered in the Base scenario, we reduce the activation of relevant KNs and evaluate the reasoning accuracy afterward. (4) Random: To establish a baseline for comparison with conditions (2) and (3), we randomly select an equal number of neurons and enhance or suppress their activation accordingly, facilitating a comparative analysis.

\paragraph{Metrics} We design a novel metric, termed \textbf{Enhance Ratio (ER)},  which serves to quantify the impact of factual retrieval failures on reasoning outcomes. ER is calculated by calculating the percentage of reasoning instances that are initially incorrect but are successfully resolved following the enhancement of KNs as Equation \ref{equation7}. Analogously, we define another metric \textbf{Suppress Ratio (SR)} to measure the obstructive effect of suppressed KNs on the reasoning process. The SR is ascertained by evaluating the ratio of cases where correct reasoning is converted to incorrect after the suppression of KNs, as outlined in Equation \ref{equation8}:
\begin{equation}
\begin{minipage}{0.88\linewidth}
\resizebox{\linewidth}{!}{$\displaystyle
\operatorname{ER} = \frac{\mathbf{|}\{\zeta \mid \operatorname{F}_{{\theta'}}(\operatorname{QT}_{r_2}(r_1(s)))= o_2\}\mathbf{|}}{|\Omega_{F}|}, \forall \zeta \in \Omega_{F} \label{equation7}
$}
\end{minipage}
\end{equation}
\begin{equation}
\begin{minipage}{0.88\linewidth}
\resizebox{\linewidth}{!}{$\displaystyle
\operatorname{SR} = \frac{\mathbf{|}\{\zeta \mid \operatorname{F}_{{\theta''}}(\operatorname{QT}_{r_2}(r_1(s))) \ne o_2\}\mathbf{|}}{|\Omega_{T}|}, \forall \zeta \in \Omega_{T} \label{equation8}
$}
\end{minipage}
\end{equation}
where ${\theta'}$ denotes the parameters of the enhanced model while ${\theta''}$ represents the parameters of the suppressed model. $QT_{r_2}(r_1(s))$ represents the reasoning question derived from two-hop fact triplets $((s,r_1,o_1),(o_1,r_2,o_2))$ with the ground truth $o_2$.
\subsection{Results}
\begin{table}[t]
    \centering
    \small
    \begin{tabular}{>{\raggedright\arraybackslash}p{0.3cm} >{\centering\arraybackslash}p{.9cm} >{\centering\arraybackslash}p{.6cm} >{\centering\arraybackslash}p{.9cm} >{\centering\arraybackslash}p{.6cm} >{\centering\arraybackslash}p{.9cm} >{\centering\arraybackslash}p{.6cm}}
    \toprule
     & \multicolumn{2}{c}{\textbf{Mistral-7B}} & \multicolumn{2}{c}{\textbf{LLaMA2-7B}} & \multicolumn{2}{c}{\textbf{LLaMA3-8B}} \\  
    \midrule
$\operatorname{Base}$ & $64.09$ & -- & $47.48$ & -- & $69.03$ & --\\
    % \cmidrule(lr){3-8}
    \midrule
    \midrule
    \textbf{Enha.}& \textbf{$\Delta$} & \textbf{ER} & \textbf{$\Delta$} & \textbf{ER} & \textbf{$\Delta$} & \textbf{ER} \\
    \midrule
    \midrule
     $\omega_1$  & 3.92  & 18.19  & 8.79 & 19.58 & 4.48 & 21.24 \\
     $\omega_2$ & 6.16  & 28.57 & 13.15 & 30.39 & 7.28 & 34.51 \\
     $\omega_{12}$  & 15.11 & \textbf{\underline{31.05}} & 15.30 & \textbf{\underline{34.97}} & 8.02 & \textbf{\underline{38.05}} \\
     $\omega_{r}$   & 4.57 & 2.74 & 7.65 & 17.79 & 0.19 & 0.88 \\
    \midrule
    \midrule
    \textbf{Supp.}& \textbf{$\Delta$} & \textbf{SR} & \textbf{$\Delta$} & \textbf{SR} & \textbf{$\Delta$} & \textbf{SR} \\
    \midrule
    \midrule
     $\omega_{1}$  & -20.06 & 32.28 & -18.00 & 38.07 & -24.53 & 38.07 \\
     $\omega_{2}$  & -29.01 & 46.70 & -24.35 & 50.78 & -39.18 & 53.03 \\
     $\omega_{12}$  & -49.53 & \textbf{\underline{77.29}} & -30.32 & \textbf{\underline{63.85}} & -62.59 & \textbf{\underline{91.61}} \\
     $\omega_{r}$  & -5.78 & 9.02 & -12.12 & 25.54 & -2.52 & 3.65 \\
    \bottomrule
    \end{tabular}
    \caption{Results of the controlled experiments after interventions on $\omega_1$, $\omega_2$ and $\omega_{12}$ under no CoT setting. $\Delta$ denotes variation in accuracy and $\omega_{r}$ is established as the baseline for enhancing or suppressing KNs of both facts, with ER/SR values expressed as percentages.}
    \label{table2}
    \vspace{-.3cm}
\end{table}
\begin{table}[t]
    \centering
    \small
    \begin{tabular}{>{\raggedright\arraybackslash}p{1.23cm} >{\centering\arraybackslash}p{.4cm} >{\centering\arraybackslash}p{.4cm} >{\centering\arraybackslash}p{.4cm} >{\centering\arraybackslash}p{.4cm} >{\centering\arraybackslash}p{.4cm} >{\centering\arraybackslash}p{.4cm}}
    \toprule
    \textbf{Models} & \multicolumn{2}{c}{\textbf{Mistral-7B}} & \multicolumn{2}{c}{\textbf{LLaMA2-7B}} & \multicolumn{2}{c}{\textbf{LLaMA3-8B}} \\
    % \cmidrule(lr){2-3} \cmidrule(lr){4-5} \cmidrule(lr){6-7}
    \midrule
    \midrule
    \textbf{Enha.}& $\omega_{base}$ & $\omega_{12}$ & $\omega_{base}$ & $\omega_{12}$ & $\omega_{base}$ & $\omega_{12}$ \\
    \midrule
    \midrule
    No CoT  & 2.74  & 31.05  & 17.79 &34.97 & 0.88 & 38.05 \\
    % \cmidrule(lr){3-8}
    Zero-shot  & 7.44  & 53.50 & 23.36 & 56.23 & 23.97 & 54.79 \\
   Few-shot  & 2.92 & 39.60 & 12.51 & 48.09 & 2.04 & 51.02 \\
   \midrule
   \midrule
   \textbf{Supp.}& $\omega_{base}$ & $\omega_{12}$ & $\omega_{base}$ & $\omega_{12}$ & $\omega_{base}$ & $\omega_{12}$ \\
    \midrule
    \midrule
    No CoT  & 9.02  & 77.29  & 25.54 & 63.85 & 3.65& 91.61 \\
    % \cmidrule(lr){3-8}
    Zero-shot   & 9.76 & 68.43& 10.80 & 71.80 & 8.25 & 74.16 \\
    Few-shot   & 0.11 & 50.48 & 5.33 & 32.09 & 0.21 & 65.92 \\
    \bottomrule
    \end{tabular}
    \caption{ ER/SR Results of enhancing and suppressing the expression of both triplets under both CoT and no CoT conditions. In the enhancement scenario, the numbers represent ER metrics, whereas in the suppression scenario, they denote SR metrics.}
    \label{table3}
    \vspace{-.3cm}
\end{table}
\paragraph{Finding 1}
In Table \ref{table2}, more than one-third of reasoning failures are caused by issues of factual retrieval. The ER values show a consistent and progressive increase as the interventions progress from targeting $\omega_{1}$, to KNs associated with the second-hop $\omega_{2}$, and ultimately to a combined intervention on both, $\omega_{12}$. This pattern indicates that many initially incorrect answers stem from retrieval failure of either the first hop, the second hop, or both during the reasoning process. Additionally, recalling the second-hop facts is more challenging for LLMs, as shown by the higher ER after enhancing $\omega_{2}$ compared to $\omega_{1}$. Suppressing factual information significantly harms reasoning performance, with accuracy dropping by over 77\%  on average when both factual elements are suppressed. Therefore, the successful retrieval of factual associations at each reasoning step is crucial for correct reasoning. 
\paragraph{Finding 2}
CoT strengthens a passive internal retrieval of relevant facts, implicitly prompting the expression of factual triplets. Evidence 1: In Table \ref{table3}, across the scenarios of no CoT, zero-shot CoT, and few-shot CoT, suppression of factual KNs results in $SR_{No\_cot}$ > $SR_{Zero\_shot}$ and $SR_{No\_cot}$ > $SR_{Few\_shot}$, which indicates that CoT likely stimulates the hydra effect \cite{mcgrath2023hydra}, which implements actively self-repairing computations to compensate the suppression effects caused by low activation levels of KNs. Evidence 2: Similarly, enhancement of factual KNs results in $ER_{No\_ cot}$ < $ER_{Zero\_shot}$ and $ER_{No\_cot}$ < $ER_{Few\_shot}$, which suggests that CoT further stimulates the internal recall process within LLMs, thus strengthening the enhancement effects of KNs. Therefore, CoT indeed can contribute to the recalling process.
% \emph{The successful retrieval of factual associations at each reasoning step is essential for achieving correct reasoning outcomes.} The ER values show a consistent and progressive increase as the interventions progress from targeting $\omega_{1}$, to KNs associated with the second-hop $\omega_{2}$, and ultimately to a combined intervention on both, $\omega_{12}$.
% This pattern indicates that a significant number of questions, which were initially answered incorrectly, were corrected to yield accurate responses following the enhancement of the corresponding KNs, as reflected by the elevated ER values. This finding highlights that a substantial portion (over 37 percent on average) of reasoning errors can be attributed to the failure in retrieving factual knowledge during the reasoning process. Furthermore, recalling the second-hop fact presents a greater challenge for LLMs, as evidenced by a higher ER after enhancing $\omega_{2}$ compared to the ER of enhancing $\omega_{1}$. Conversely, the suppression of factual information has a pronounced negative impact on reasoning performance. Notably, when the activation values of KNs related to two-hop questions are nullified, there is a significant drop in accuracy, with a decline exceeding 77 percent on average following the suppression of the recall of both factual elements.
\begin{figure*}[!t]
	\centering
        \subfigure{\includegraphics[width=.328\linewidth]{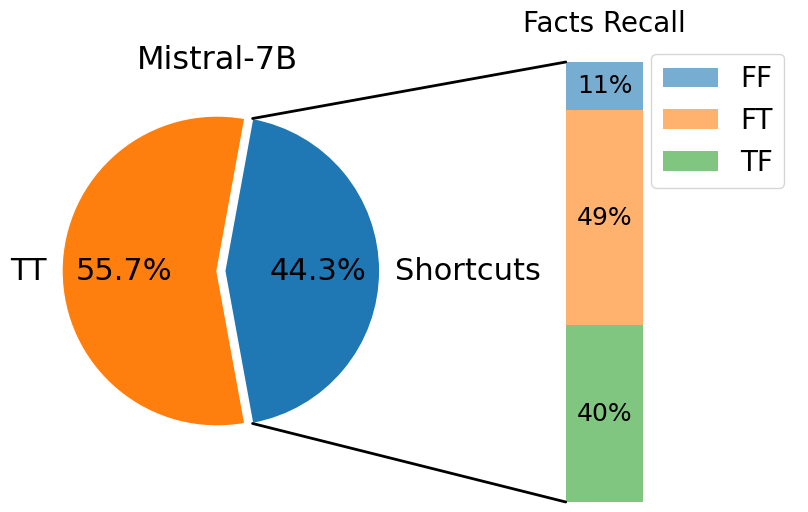}}
	\subfigure{\includegraphics[width=.328\linewidth]{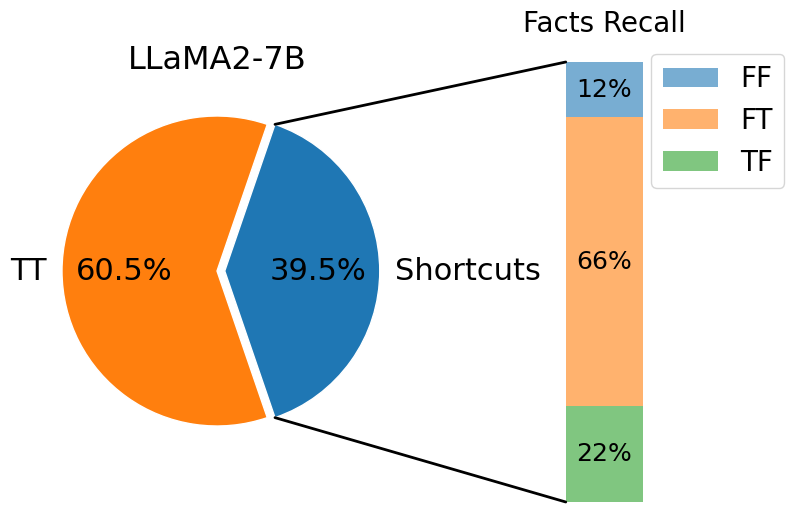}}
    \subfigure{\includegraphics[width=.328\linewidth]{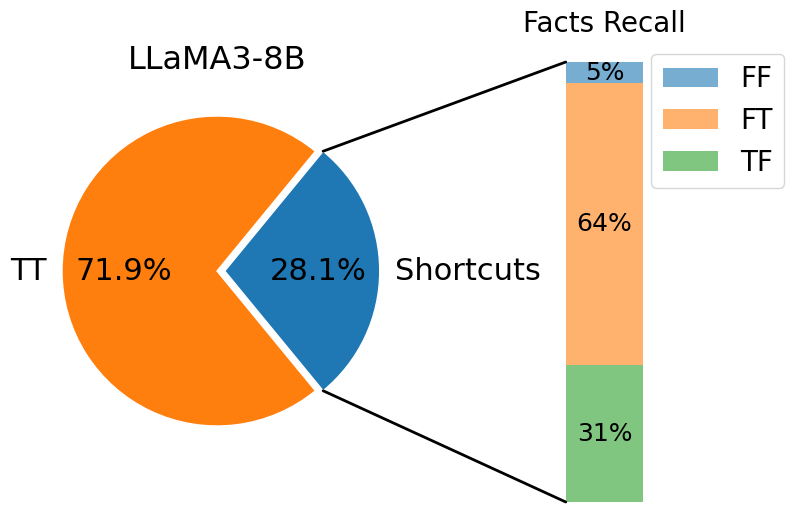}}
    % \vspace{-.1cm}
	\caption{An in-depth analysis of shortcut scenarios under no CoT. TT represents successful recall of both facts.}
	\label{fig4}
 % \vspace{-.3cm}
\end{figure*}
\section{Analysis of Shortcuts}
% In both sections 4 and 5, we explore whether improving the recall of factual knowledge in the reasoning process could contribute to the success of two-hop reasoning. Conversely 
In this section, we investigate whether successful two-hop reasoning implies the successful recall of factual knowledge. In other words, we examine whether accurate reasoning outcomes stem from a thorough grounding in multi-hop knowledge reasoning or are facilitated by alternative shortcuts.
\subsection{Experiment}
\paragraph{Setup}We investigate the utilization of individual fact triplets in correctly answered two-hop questions by analyzing the KN Scores for each triplet. We compare these scores with those observed during single-hop queries to establish a threshold, denoted as $\tau$, which serves as a benchmark for identifying the effective use of facts in the reasoning process. If the activation level of KNs falls significantly below this threshold in comparison to single-hop queries, this indicates an under-utilization of the corresponding fact. Conversely, if it exceeds the threshold, the fact is considered adequately utilized. Using this criterion, we classified the correctly answered questions into four distinct categories: (1) \textbf{FT}: Unsuccessful recall of the first-hop fact but successful second-hop recall; (2) \textbf{TF}: Successful first-hop recall but unsuccessful second-hop recall; (3) \textbf{FF}: Neither fact successfully recalled and (4) \textbf{TT}: Both facts successfully recalled. Except for TT, the other three situations are defined as \emph{Shortcuts}.
% We investigate the utilization of individual fact triplets in correctly answered two-hop questions by analyzing the KN Scores for each triplet. We compared these levels against those observed during single-hop queries to define a threshold, denoted as $\tau$, which serves as a benchmark for identifying the effective use of facts in the reasoning process. If the activation level of KNs falls notably below this threshold, particularly more so than those under a single-hop query, it indicates an under-utilization of the corresponding fact; conversely, if it exceeds the threshold, the fact is considered to have been adequately utilized. Employing this criterion, we classified the correctly answered questions into four distinct categories: (1) \textbf{FT}: those where the recall of the first-hop fact was unsuccessful but the second-hop fact was successfully recalled; (2) \textbf{TF}: those with a successful recall of the first-hop fact but an unsuccessful recall of the second-hop fact; (3) \textbf{FF}: those where both facts were unsuccessfully recalled; and (4) \textbf{TT}: those with successful recall of both facts. Except TT, three other situations are defined as \emph{Shortcuts.}
\begin{table}[ht]
    \small
    \centering
    \begin{tabular}{l p{.5cm} p{.5cm} p{.5cm} p{.5cm} p{.5cm} p{.5cm}}
    \toprule
    \textbf{Models} & \multicolumn{2}{c}{\textbf{Mistral-7B}} & \multicolumn{2}{c}{\textbf{LLaMA2-7B}} & \multicolumn{2}{c}{\textbf{LLaMA3-8B}} \\
    \cmidrule(lr){2-3} \cmidrule(lr){4-5} \cmidrule(lr){6-7}
    & MH & SC & MH & SC & MH & SC \\
    \midrule
     No CoT  & 55.75  & 44.25  & 60.51 & 39.49 & 71.89 & 28.11 \\
     Zero-shot  & 70.84  & 29.16 & 64.26 & 35.74 & 95.66 & 4.34 \\
     Few-shot  & \textbf{91.23} & \textbf{\textcolor{red}{8.77}} & \textbf{89.02} & \textbf{\textcolor{red}{10.98}}& \textbf{97.65} & \textbf{\textcolor{red}{2.35}} \\
    \bottomrule
    \end{tabular}
    \caption{The fraction of MH and SC in correctly answered examples (TT+FT). MH denotes successful retrieval of both facts while others denote by SC.}
    \label{table5}
    \vspace{-.5cm}
\end{table}
\subsection{Results Analysis} 
According to Table \ref{table5}, under normal conditions, a considerable proportion of correctly answered questions under no CoT setting rely on shortcuts, possibly due to word associations intrinsic to LLMs,as observed by \citet{yang2024large}. Notably, the Mistral-7B model stands out for its unexpected reliance on shortcuts to solve over 44 percent of the questions successfully. Even with large-scale models possessing 7 billion parameters, LLMs still rely on certain segments of the reasoning chain to arrive at answers.
The introduction of CoT effectively decreases the trend of taking shortcuts by forcing LLMs to recall more relevant facts and engage in multi-hop reasoning. Under few-shot CoT setting, all LLMs solve over 90 percent of questions on average through multi-hop reasoning, reducing the ratio of shortcuts to nearly zero.

Figure \ref{fig4} provides a closer look at the shortcut phenomenon. The percentage of FF is significantly low, illustrating that it is hard for LLMs to fail to retrieve any factual information relevant when presented with the clues of overlapping entities or relational vocabulary in queries. For most instances of shortcuts, LLMs prefer to utilize the second-hop fact to directly answer reasoning questions, skipping the intermediate reasoning steps and relying on the object $o_2$ in the second-hop to cheat (a high ratio for FT). For TF cases, there might exist direct associations between the head entity $s$ and the tail entity $o_2$ leveraged to derive correct answers. In conclusion, experimental results show that recalling two-hop facts (TT) benefits the model's reasoning performance. Specifically, with the presence of CoT, the proportion of TT significantly increases and the model's reasoning accuracy improves substantially.
\section{Impact of Contextual Conflict}
The capacity of utilizing internal factual knowledge is contingent not solely upon the intrinsic properties of LLMs, but is also significantly influenced by the context within which they operate. This section elucidates how the presence of knowledge conflicts within a given context can impact the mechanisms of the retrieval process during reasoning.
\begin{table*}[ht]
    \centering
    \small
    \begin{tabular}{ l l l}
    \toprule
   \textbf{Relational Facts} & \textbf{Type} & {\textbf{Examples}} \\
   \midrule
 \multirow{6}{*}{\tabincell{l}{$\langle$ \texttt{Middlemarch,} \\ ~~\texttt{author,}\\  ~~\texttt{George Eliot}$\rangle$\\ \\$\langle$\texttt{George Eliot,}\\ ~~\texttt{place of death,} \\ ~~\texttt{London}$\rangle$}}& \multirow{2}{*}{Distraction} & Context: Carl Sagan works at Cornell University. \\
    &&Question: Where did the author of Middlemarch pass away? A:\\
    \cmidrule{2-3}
    & \multirow{2}{*}{Conflict 1} & Context: The author of Middlemarch is Jean Genet.\\
    && Question: Where did the author of Middlemarch pass away? A:\\
    \cmidrule{2-3}
    & \multirow{2}{*}{Conflict 2} & Context: George Eliot died in the city of Atlanta.\\
    && Question: Where did the author of Middlemarch pass away? A:\\
    \bottomrule
    \end{tabular}
    % \vspace{-.1cm}
    \caption{Knowledge conflict and knowledge distraction examples}
    \label{table4}
    % \vspace{-.3cm}
\end{table*}
\subsection{Experiment}
\paragraph{Setup} For each data point, we formulate a single-hop conflict fact by devising a set of potential objects denoted as $O_{candi}$ for its $r$. From this set, we deliberately select an object $o^{*} \ne o$ to introduce a knowledge conflict. In contradistinction, we also fabricate an entirely unrelated fact for each data point to serve as a distractor, referred to as knowledge distraction (See detailed construction in Appendix \ref{D}). We then respectively append the knowledge conflict and knowledge distraction sentences before the two-hop question under no CoT setting, which is input into LLMs. Then we observe the values of KN Scores for each-hop fact. The examples of knowledge conflict and distraction for the first-hop and the second-hop facts are shown in Table \ref{table4}.
\begin{figure}[!t]
    \centering
    \includegraphics[width=.99\linewidth]{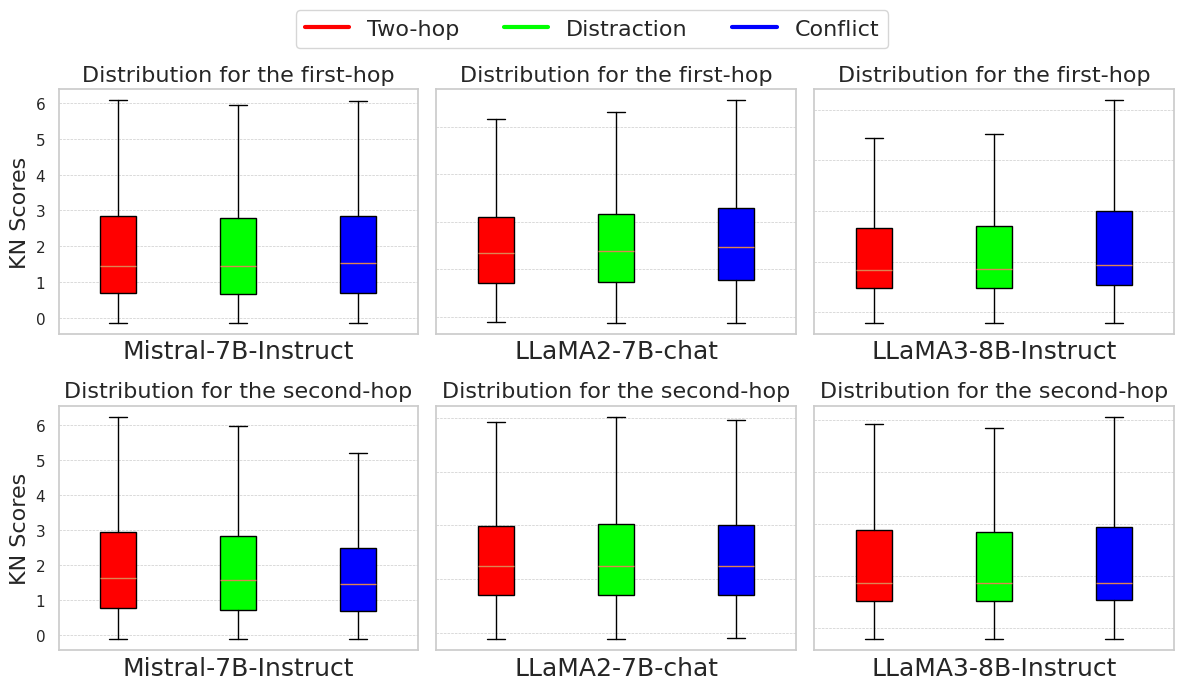}
    \small
    \caption{Results of constructing the knowledge distraction and knowledge conflict for the first-hop fact.}
    \label{fig5}
    \vspace{-.5cm}
\end{figure}
\subsection{Results Analysis}
\begin{figure}[!t]
    \centering
    \includegraphics[width=.99\linewidth]{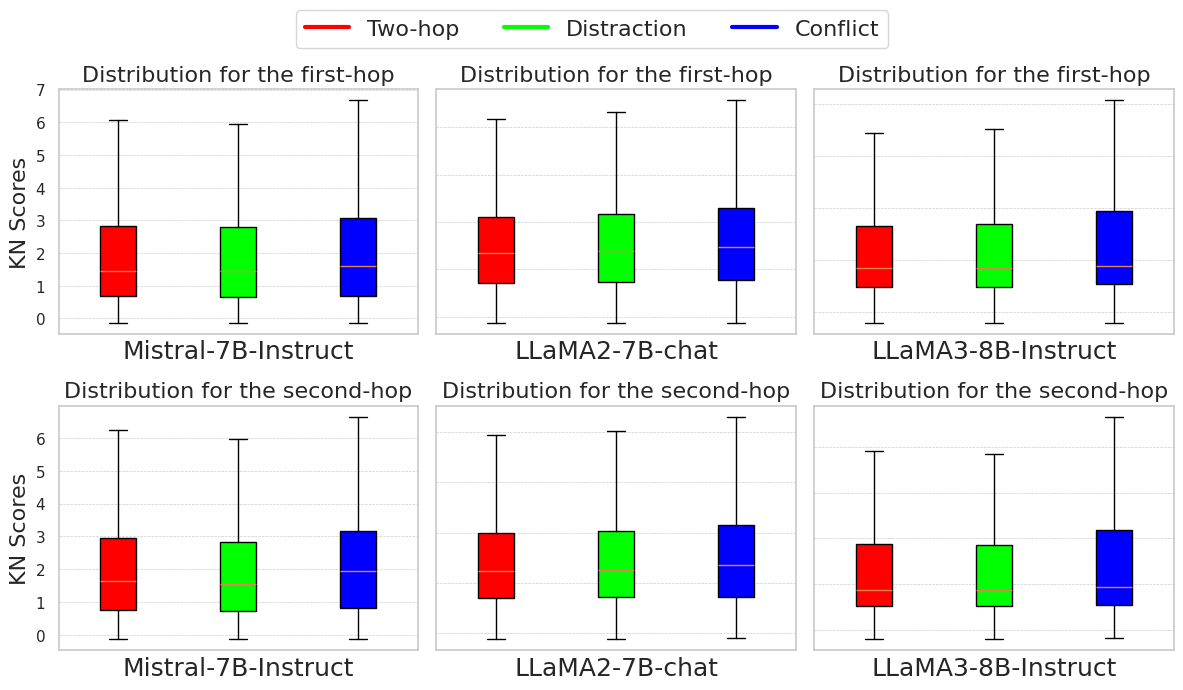}
    \small
    \caption{Results of constructing the knowledge distraction and knowledge conflict for the second-hop fact.}
    \label{fig6}
    \vspace{-.3cm}
\end{figure}

The presence of knowledge conflict within the context consistently augments the faithfulness of LLMs in the corresponding fact. According to Figure \ref{fig5} and Figure \ref{fig6}, the context of knowledge conflict results in the highest KN Scores of the corresponding hop fact \footnote{The results were obtained from a one-tailed paired sample t-test, conducted at a significance level of 0.05.}, which indicates counterfactual context significantly improves the internal retrieval of that corresponding hop fact. It illustrates LLMs exhibit greater confidence in their encoded knowledge when confronted with knowledge conflict, a finding that aligns with the studies conducted by \citet{zhou-etal-2023-context} and \citet{li-etal-2023-large}. 
When the knowledge presented in the context conflicts with the second-hop fact, it not only reinforces the retrieval of the second-hop fact but also enhances the recall of the first-hop fact. It is plausible that the introduction of the subject $o_1$ encourages LLMs to recall the precise triplet $(s, r_1, o_1)$. However, this effect does not extend to the first-hop fact. The occurrence of knowledge distraction appears 
 not to cause much obstruction to the factual recall within LLMs. On the contrary, it may even stimulate LLMs to retrieve more facts sometimes, as evidenced by the high KN Scores for the first-hop fact of LLaMA2-7B when the knowledge distractor corresponding to the second-hop fact appears in Figure \ref{fig6}.
\section{Related Work}
\paragraph{Multi-hop Reasoning} Multi-hop reasoning poses a significant challenge for LLMs. Several studies have endeavored to address this challenge through the development of more faithful reasoning techniques \cite{creswell2022faithful,NEURIPS2023_c518f504,creswell2023selectioninference}. One such approach is CoT, which stimulates LLMs to produce deductive intermediate steps, fostering a step-by-step analytical process \cite{chu2024navigate}. Another line of research is focused on visualizing the implicit logical structures within LLMs from the perspective of mechanistic interpretability \cite{yang2024large}. For example,  a recent study by \citet{hou-etal-2023-towards} recovers the reasoning tree from models’s attention patterns using MechanisticProbe.
% Recent studies have demonstrated the remarkable performance of language models across various reasoning tasks. However, the underlying reasoning mechanisms of these models remain unclear—whether they simply memorize superficial data relations from pretraining corpora or perform robust reasoning processes. Unveiling the internal mechanisms of model reasoning not only enhances model interpretability but also provides clues for building trustworthy language model reasoners. In prior work, a line of research has turned to mechanistic interpretability \cite{merullo2024language,wu2023interpretability,bayazit2023discovering,dutta2024think,chen2024da,chen2024journey}. For example, \citet{hou-etal-2023-towards} proposes a novel probing approach MechanisticProbe, to recover the model's internal reasoning trees from attention patterns, for multi-step reasoning tasks.  
\paragraph{CoT Mechanism} A large body of literature is dedicated to the theoretical and empirical exploration of the mechanism underlying CoT 
 \cite{saparov2023language,tan-2023-causal,feng2023towards,prystawski2023why,xie2024calibrating}. Some research endeavors to delve into a reverse-engineering analysis of CoT prompting, uncovering the intricate information pathways that facilitate the generation of responses \cite{dutta2024think}. However, the majority of these studies concentrate on the rationales produced by CoT and have largely overlooked the broader implications for factual retrieval processes. In our current work, we complement this aspect and present compelling evidence that CoT significantly bolsters the internal recall of factual information.
 % by encouraging LLMs to undertake a meticulous, step-by-step reasoning approach, thereby reducing the propensity for these models to resort to shortcuts.
 % and how they were written to the output residual stream through probing attention heads
  % To circumvent the MLP blocks associated with factual memory \cite{geva-etal-2021-transformer}, they strategically use reasoning problems that involve fictional entities, thereby bypassing the reliance on factual knowledge inference within LLMs. Our research complements existing studies by examining the impact of CoT on the retrieval of factual memories within LLMs.
% By applying activation patching\cite{wang2022interpretability}, they pinpoint the information pathways that process the answer and how it was written to the output residual stream through probing attention heads. 
\section{Conclusions} This paper aims to provide a comprehensive understanding of factual recall behaviors for LLMs. We find that a considerable portion of reasoning failures are due to retrieval failures. Manually enhancing the internal recall within LLMs can improve reasoning performance. For LLMs, they not only rely on multi-hop reasoning but also rely on other inference ways in LLMs such as shortcuts. CoT can significantly stimulate LLMs to recall more facts by compelling models to engage in step-by-step thinking, diminishing the possibilities of taking shortcuts. The knowledge conflict existing in context could improve the confidence of parametric knowledge, therefore enhancing the internal recall.

% On the one hand, LLMs do not proactively extract and apply their internal knowledge for reasoning during the reasoning process. However, the enhanced recall of pertinent facts in two-hop reasoning questions substantially improves multi-hop performance. On the other hand, for questions answered correctly, LLMs rely on shortcuts in addition to grounded multi-hop reasoning. Particularly without CoT, it is challenging for LLMs to avoid shortcuts. CoT compels LLMs to engage in systematic thinking and to perform multi-hop reasoning, thereby significantly reducing the likelihood of taking shortcuts. Precisely because of this engagement in systematic thinking, LLMs are encouraged to enhance the internal recall of factual knowledge essential for reasoning. LLMs often fail to fully harness the knowledge they possess; however, retrieval enhancement can somewhat compensate for this shortcoming.
\section*{Limitations}
While our study provides novel insights into the internal factual recall behaviors of LLMs during reasoning tasks, it is important to acknowledge several limitations. 
\paragraph{Generalizability:} While the current study is primarily based on specific LLMs and the TFRKN dataset, future research should extend these findings to verify their generalizability across various models and datasets
\paragraph{Theoretical Analysis:} Although empirical evidence has been provided through targeted interventions, a deeper theoretical analysis is needed to fully comprehend the underlying reasons for the observed phenomena.
\paragraph{Practical Applications:} The paper discusses theoretical aspects and potential improvements in reasoning accuracy but does not delve into how these findings can be applied in practical scenarios to enhance the reasoning capabilities of LLMs.
\paragraph{Impact of Contextual Factors:}While the paper touches upon the influence of contextual conflicts on knowledge retrieval, a more comprehensive analysis of various contextual factors and their impact on reasoning is needed.
\section*{Acknowledgements}
This work was supported in part by the Strategic Priority Research Program of Chinese Academy of Sciences under Grant \#XDA27030100 and the National Natural Science Foundation of China under Grants \#72293573.
\bibliography{custom}

\appendix
\section{Details of Dataset Construction }
\label{A}

\subsection{Sampling two-hop factual triples}
Our dataset is constructed based on Wikidata \cite{vrandevcic2014wikidata}, a structurally optimized database covering nearly all domains. The dataset is available at \url{https://github.com/wangyifei0047/TFRKN}. 

First we show \textbf{manually selected relations} that are used to construct two-hop relations: 
\begin{itemize}[itemsep=1pt, parsep=0pt, topsep=0pt,leftmargin=15pt]
    \item P30, P36, P35, P1037, 1308, P164, P449, P488, P178, P159, P286, P413, P641, P800, P937
    \item P136, P106, P495, P740, P37, P407, P170, P50,P364,P112, P108, P175, P27, P40, P69, P19
\end{itemize}
While LLMs have been shown to store a vast amount of factual knowledge, studies indicate that they are more likely to recall triplets related to popular entities \cite{mallen-etal-2023-trust}. Therefore, when constructing the dataset, we employ the cumulative pageviews count over the past 12 months as a measure and select the top 500 popular entities based on this criterion. Two-hop reasoning chains are then extracted from the sub-graphs consisting solely of the aforementioned relations and entities, like \emph{(Holden Caprice, manufacturer, General Motors), (General Motors, chairperson, Mary Barra)}.
\subsection{Generating Queries using ChatGPT}
\label{A.2}
\begin{figure*}[t]
\centering
    % \vspace{-0.5in} % 调整这个值以控制图像与页面顶部的距离
    \includegraphics[width=0.98\textwidth]{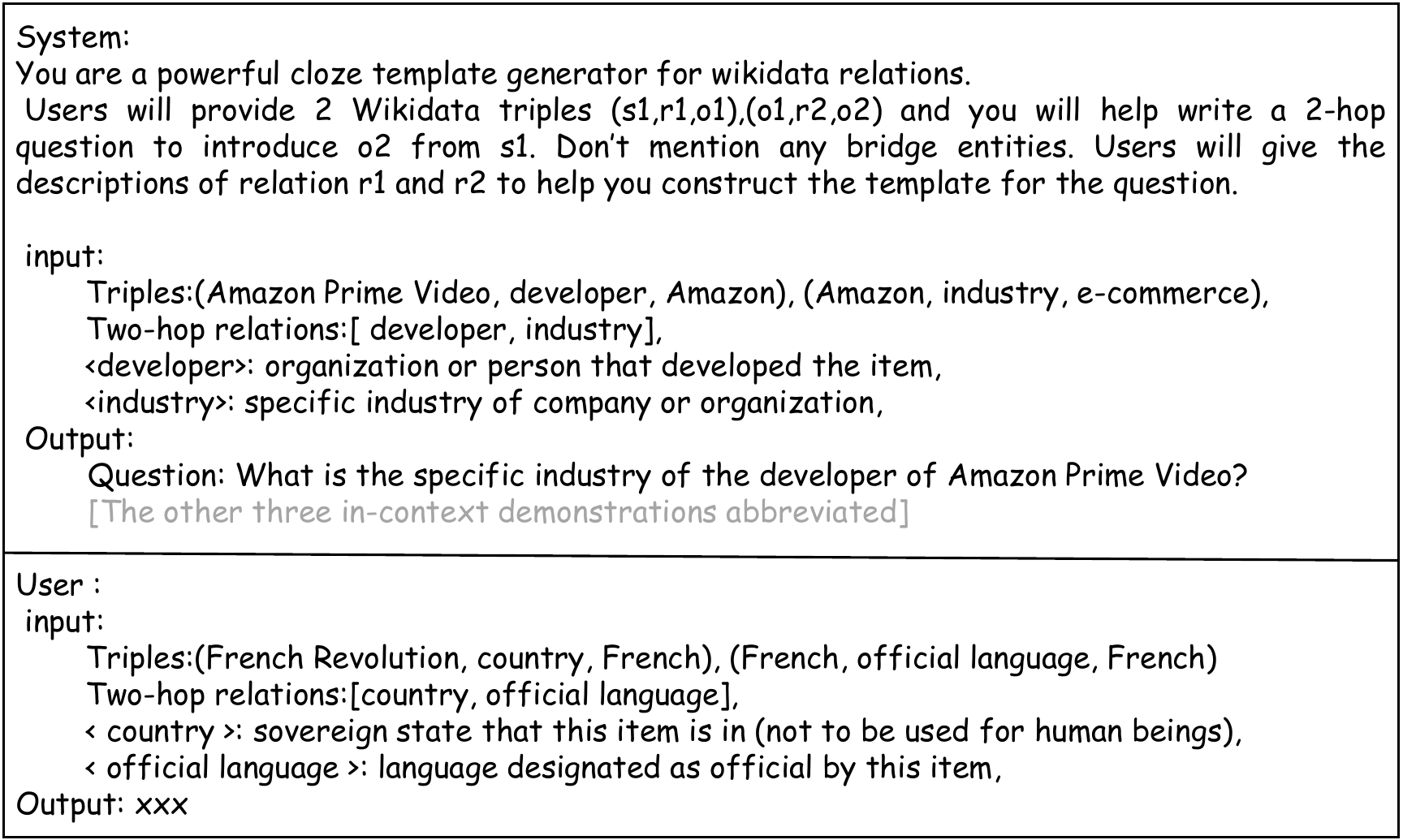}
    \caption{An example of using ChatGPT to generate 2-hop questions from Wikidata triples.}
    \label{fig7}
\end{figure*}

\begin{figure*}[!h]
    \centering
    \includegraphics[width=0.98\textwidth]{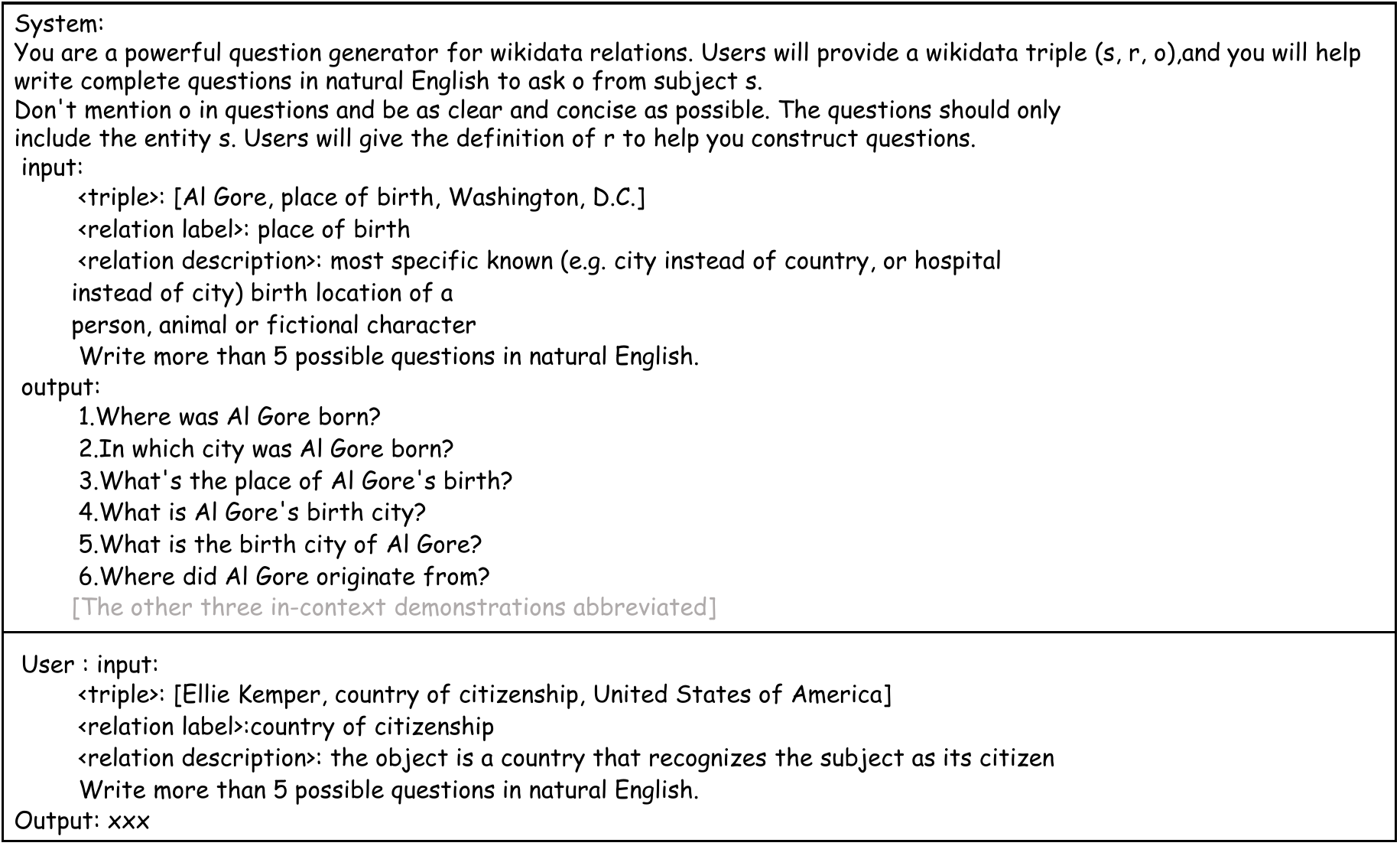}
    \caption{An example of using ChatGPT to generate single-fact queries from triples and relation information(labels and descriptions).}
    \label{fig8}
\end{figure*}

 Having acquired the triplet format of reasoning queries, our current objective is to transform these triplets into natural language expressions in queries. Moreover, for effective integration of the Knowledge Neuron technique, it is essential to rephrase individual triplets into multiple natural language expressions. As knowledge neurons demonstrate indifference towards specific knowledge representations, employing diverse question formats aids in identifying authentic knowledge neurons. Whether in the formulation of reasoning queries or the generation of individual triplet queries, we capitalize few-shot learning capabilities of ChatGPT (gpt-3.5-turbo) to autonomously generate natural language questions. Concretely, we leveraged few-shot capabilities in LLMs to generate multiple queries for individual fact $(s,r,o)$, as well as reasoning questions from two-hop facts $((s_1,r_1,o_1),(o_1,r_2,o_2))$. For the generation of single-fact queries, we provide relation labels and relation definitions as additional information for LLMs to generate accurate subject-relation queries (Figure \ref{fig8}). For the generation of reasoning questions, two-hop relation labels and explanations are also provided besides four in-context demonstrations (Figure \ref{fig7}). 
 
 An instance from TFRKN is depicted in Table \ref{table6}. This approach not only surpasses the limitations imposed by manual templates but also guarantees the production of high-quality and diverse questions. Overall, the dataset comprises 4,550 instances spanning 213 unique combinations of relations.
\begin{table}[h]
\centering
\renewcommand{\arraystretch}{1.1}
\footnotesize
\begin{tabular}{>{\raggedright\arraybackslash}p{1.5cm} >{\raggedright\arraybackslash}p{5.5cm}}
\toprule
\multirow{2}{*}{$\operatorname{Triples}$} & (Holden Caprice, manufacturer, General Motors)\\  
                & (General Motors, chairperson, Mary Barra) \\
\midrule
\multirow{5}{*}{$\operatorname{Fact_1 Query}$}     & 1. Who or what company manufactures Holden Caprice?\\
              & 2. What company created Holden Caprice?\\
              & 3. Who is responsible for making Holden Caprice?\\
              & 4. What entity produces Holden Caprice?\\
              & 5. Which organization is behind the production of Holden Caprice?\\
\midrule
\multirow{5}{*}{$\operatorname{Fact_2 Query}$}        & 1. Who is the chairperson of General Motors?\\
              & 2. Who is the head of General Motors?\\
              & 3. Who presides over General Motors as its chairperson?\\
              & 4. Who currently serves as the chairperson of General Motors?\\
              & 5. What is the name of the person who chairs General Motors?\\
\midrule
$\operatorname{Reason\_Q}$ & Who is the chairperson of the manufacturer of Holden Caprice?\\
\bottomrule
\end{tabular}
\caption{An instance from TFRKN}
\label{table6}
\vspace{-.6cm}
\end{table}
\section{Knowledge Neurons}
\label{B}
In this part, we detailedly illustrate the methodology of the identification of KNs using the integrated gradient method. Given a specific relational fact: $(s, r, o)$; A set of knowledge-expressing queries ( Fact1Query and Fact2Query in Table \ref{table6}): $<query_1, query_2, \cdots, query_L>$. We define the representation of the $i$-th neuron in the $l$-th intermediate layer in FFNs as $w_i^l$,
\begin{equation}
\small
\begin{aligned}
    P_{\left[{t_1, \cdots, t_n} \right],y}(w_i^{(l)})= P(y| \left[{t_1,\cdots, t_n} \right],w_i^{(l)}=\overset{\sim}{w}_i^{(l)})
\end{aligned}
\label{equation9}
\end{equation}
where $\left[{t_1, t_2, \cdots, t_n} \right]$ represents the token sequence of inputs, $\overset{\sim}{w}_i^{(l)}$ represents the constant value assigned to $w_i^{(l)}$, and Equation \ref{equation9} denotes the probability of next token y predicted by LLMs, given the token sequence $\left[{t_1, t_2, \cdots, t_n} \right]$ after $w_i^{(l)}$ is assigned the value $\overset{\sim}{w}_i^{(l)}$.\\
The attribution scores quantify the contribution of individual neurons to correct predictions. By gradually restoring each neuron's value from 0 to its original level, the gradients of the probability of the correct token with respect to each neuron are integrated, as shown in Equation \ref{equation10}.\\
Equation \ref{equation10} is applied to the calculation of attribution scores for single-token target \( o \). The method for computing attribution scores for multi-token target \( o \) is described in Equation \ref{equation11}. Assuming the tokenized sequence of a relational-fact query and the corresponding ground truth respectively are $\left [q_1, q_2, \cdots, q_n\right ]$ and $\left [ gt_1, gt_2, \cdots, gt_m \right ]$.
\begin{equation}
\small
Attr(w_i^{(l)})=\overline{w}_i^{(l)}\int_{\beta=0}^{1}\frac{\mathrm{d} P_{\left[{t_1, \cdots, t_n} \right],y} (\beta
 \overline{w}_i^{(l)})}{\mathrm{d} w_i^{(l)}} {\mathrm{d}\beta}
 \label{equation10}
\end{equation}
\begin{equation}
\small
\begin{aligned}
    &\overset{\sim}{\text{Attr}}(query, w_{i}^{(l)})=\\ 
    &\frac{1}{m}\sum_{k=1}^{m} \overline{w}_{i,k}^{(l)} \int_{\beta=0}^{1}\frac{\mathrm{d} P_{\left[{q_1, \cdots, q_n,\cdots, a_{k-1}} \right],gt_k} (\beta  \overline{w}_{i,k}^{(l)})}{\mathrm{d} w_{i,k}^{(l)}}{\mathrm{d}\beta}
\end{aligned}
\label{equation11}
\end{equation}
where $a_i$ represents the generated token with the highest predicted probability at $i$-th time. Due to the intractability of the continuous integration in Equation \ref{equation10}, an approximation is made using Riemann integration (equation \ref{equation12}). Substituting Equation \ref{equation12} into Equation \ref{equation11} yields Equation \ref{equation13}.
\begin{equation}
\small
    \begin{aligned}
  &Attr(w_i^{(l)}) = \frac{\overline{w}_{i}^{(l)}}{N}\sum_{j=1}^{N}\frac{\partial P_{\left[{t_1, \cdots, t_n} \right],y} (\frac{j}{N}\overline{w}_i^{(l)})}{\partial w_i^{(l)}} \label{equation12}\\
\end{aligned}
\end{equation}
\begin{equation}
\small
    \begin{aligned}
  &\overset{\sim}{Attr}(query,w_{i}^{(l)}) = \\
  &\frac{1}{m}\sum_{k=1}^{m} \frac{\overline{w}_{i,k}^{(l)}}{N} \sum_{j=1}^{N}\frac{\partial P_{\left[{q_1,\cdots, q_n,a_1, \cdots, a_{k-1}} \right],gt_k} (\frac{\overline{w}_{i,k}^{(l)})}{N}) }{\partial w_{i,k}^{(l)}}  
\end{aligned}
\label{equation13}
\end{equation}
Given that knowledge neurons surpass linguistic expressions and govern the expression of authentic knowledge, we retain knowledge neurons shared by more than $p\%$ queries as Equation \ref{equation14}.
\vspace{-.2cm}
\begin{equation}
\small
    \begin{aligned}
KN= &\underset{k=1}{\overset{L}{\bigcap}}KN_{query_k}\\
\vspace{-.1cm}
    KN_{query_k} = &\{ w_{i}^{(l)}| \overset{\sim}{Attr}(query_k,w_{i}^{(l)})> \tau, \forall i,l\}   
\end{aligned}
\label{equation14}
\end{equation}
\section{Experimental Details}
\label{C}
We present a comprehensive overview of our experimental setup. 
\paragraph{Intersection of LLMs} Experiments are conducted using a refined subset of TFRKN dataset. To ensure that LLMs know each factual element required by the factual reasoning questions, we meticulously filtered out unqualified data points for each model. By taking the intersection of these filtered datasets, we culled a dataset comprising 1072 qualified data points. 

\paragraph{Indentification of KNs} The process of identifying KNs for each fact triplet proves to be the most computationally intensive, with each model taking 96 GPU hours to find all KNs. In the context of the location experiment, we configured the integrated gradient steps to 20 and set the parameter of the shared percentage of coarse neurons to 0.2. The experiments were executed on a system equipped with NVIDIA A100 80GB GPUs, and further details of the software environment are available in our code repository. All experimental results are the mean values of three repetitive experiments.
\section{Construction of Contextual Conflict}
\label{D}
% For knowledge conflict experiments, we construct a knowledge distraction sentence pool, randomly assigned to each reasoning question while knowledge conflict in the cloze task is constructed by a set predefined templates of relations.
\paragraph{Knowledge Distraction} We manually constructed a set of irrelevant fact statements $S$. $S$ does not involve any entities or relations in TFRKN to ensure "unrelated" property. Each two-hop question randomly selects a knowledge distraction from this set.
\paragraph{Knowledge Conflict} We constructed contexts that conflict with the first-hop fact and that conflict with the second-hop fact for each two-hop question respectively. The method is as follows: we manually designed templates $T$ for all relations involved in the TFRKN dataset. Assuming there is a fact $(s,r,o)$, we collect the set of candidate objects related to $r$ in the dataset, select an $o^*$
that is not equal to $o$ as the new fabricated fact $(s,r,o^*)$, and apply the template of the 
 relation in $T$ to obtain the knowledge conflict context corresponding to $(s,r,o)$.
\section{Additional Experimental Results}
\label{E}
\begin{table*}[!t]
    \centering
    \begin{tabular}{cccccccccc}
    \toprule
    & \multicolumn{3}{c}{\textbf{LLaMA2}} & \multicolumn{3}{c}{\textbf{LLaMA3}} & \multicolumn{3}{c}{\textbf{Mistral}} \\
    \cmidrule{2-10}
    & Avg. & Med. & Max. & Avg. & Med. & Max. & Avg. & Med. & Max. \\
    \midrule
    Number of KNs per fact & 26.4 & 22.0 & 53.0 & 28.3 & 25.0 & 52.0 & 22.3 & 26.0 & 59.0 \\
    Number of pairwise intersections & 2.7 & 2.0 & 6.0 & 3.7 & 3.0 & 4.0 & 1.0 & 0.0 & 4.0 \\
    \bottomrule
    \end{tabular}
    \caption{The KNs for different facts may vary significantly (Avg.: average, Med.: median, Max.: maximum).}
\label{table7}
\end{table*}
\begin{table*}[!t]
    \centering
    \begin{tabular}{p{8cm}ccc}
    \toprule
    Sentences ending with Paris& \textbf{LLaMA2-7B} &\textbf{LLaMA3-8B} & \textbf{Mistral-7B} \\
    \midrule
     \textbf{The capital of France is} & 	\textbf{1.98}	&\textbf{2.25}	&\textbf{2.04}\\
    % \midrule
     \textbf{The capital city of France is} & 	\textbf{2.01}	& \textbf{2.27}	&\textbf{2.06}\\
     \midrule
     The Louvre Museum is situated in the city of & 	1.10	&1.46	&1.55\\
     \midrule
     The Seine River flows gracefully through the heart of & 	1.27	&1.47	&1.70\\
     \midrule
     The City of Love refers to the city of & 	1.24	&1.70	&1.41\\
     \midrule
     The City of Light refers to the city of & 	1.25	&1.65	&1.53\\
     \midrule
     The capital The football club Paris Saint-Germain is based in the city of France is & 	1.24	&1.40	&1.66\\
     \midrule
     The Eiffel Tower is one of the most iconic landmarks in the city of & 	1.37	&1.80	&1.64\\
    \bottomrule
    \end{tabular}
    \caption{KN Scores corresponding to $(France, capital, Paris)$ for different sentences which end with Paris.}
\label{table8}
\end{table*}
\paragraph{Non-overlap of Knowledge Neurons} Based on the KNs identified in our experiments, we conducted a verification of non-overlap. To achieve this, we randomly sampled 6,000 pairs of distinct relational facts and calculated the number of intersecting KNs between each pair. The statistics of overlapping KNs are shown in Table \ref{table7}.
% "The capital of France is" and "The capital city of France is" are knowledge-expressing prompts, which consistently exhibit higher KN Scores compared to other examples, even though LLMs predict "Paris" for all these sentences. 

\paragraph{Verification of Basic Assumptions} We present compelling results from small-scale case studies, which prove that when LLMs predict the same word, the metric of KN Scores would be high only when the process involves fact retrieval. For illustration, we use $(France, capital, Paris)$ as an example, whose KNs cover 26 neurons. KN Scores are computed across these 26 neurons as the metric. Then we construct sentences that end with "Paris" and then replace "Paris" with a blank, prompting LLMs to predict the missing word. To ensure that the LLMs predict "Paris" as the final token, we design straightforward and commonsense sentences and verify that the LLMs would indeed predict "Paris" and then assess the knowledge-expressing prompts and compare them with non-knowledge-expressing prompts by analyzing their KN Scores. 

"The capital of France is" and "The capital city of France is" are knowledge-expressing prompts, which consistently exhibit higher KN Scores compared to other examples, even though LLMs predict "Paris" for all these sentences. This experiment illustrates that KNs for $(France, capital, Paris)$ are activated mostly when LLMs recall $(France, capital, Paris)$, not when they make a specific predictive word "Paris".
\end{document}

%% file: settings.tex
\usepackage{multirow}
\usepackage{amsmath}
\usepackage{capt-of}
\usepackage{tabularx}
\usepackage{epsfig}
\usepackage{amssymb}
\usepackage{amsfonts}
\usepackage{booktabs}
\usepackage{scalerel}
\usepackage[inline]{enumitem}
\usepackage{listings}
\usepackage{varwidth}
\usepackage[export]{adjustbox}
\usepackage{tikz}
\usetikzlibrary{tikzmark}
\usepackage{cleveref}

\usepackage{stmaryrd}
\usepackage{bbm}

\newcommand{\tabincell}[2]{\begin{tabular}{@{}#1@{}}#2\end{tabular}}

\usepackage{algorithm}
\usepackage[noend]{algpseudocode}

\definecolor{deepblue}{rgb}{0,0,0.5}
\definecolor{officeblue}{RGB}{0,102,204}
\definecolor{deepred}{rgb}{0.6,0,0}
\definecolor{deepgreen}{rgb}{0,0.5,0}
\definecolor{mybrickred}{RGB}{182,50,28}

\definecolor{fillcolor}{RGB}{216,217,252}

\algnewcommand\algorithmicrequireb{{\hspace{0.85cm}}}
\algnewcommand\INPTDESCB{\item[\algorithmicrequireb]}

\algnewcommand\algorithmicfuncdesc{\textbf{Function:}}
\algnewcommand\FUNCDESC{\item[\algorithmicfuncdesc]}
\algnewcommand\algorithmicfuncdescb{{\hspace{1.48cm}}}
\algnewcommand\FUNCDESCB{\item[\algorithmicfuncdescb]}
\algnewcommand{\algorithmicgoto}{\textbf{goto}}
\algnewcommand{\Goto}[1]{\algorithmicgoto~\ref{#1}}

%% file: math.tex
\usepackage{amsmath,amsfonts,bm}

\def\eqref#1{equation~\ref{#1}}
\def\1{\bm{1}}

\DeclareMathAlphabet{\mathsfit}{\encodingdefault}{\sfdefault}{m}{sl}
\SetMathAlphabet{\mathsfit}{bold}{\encodingdefault}{\sfdefault}{bx}{n}